\documentclass{article}
\usepackage{spconf,amsmath,graphicx,hyperref, amssymb,booktabs,multirow}

\title{Long Chain-of-Thought Compression via Fine-Grained Group Policy Optimization}

\name{Xinchen Han$^{1}$ \qquad Hossam Afifi $^{1}$ \qquad Michel Marot$^{1}$ \qquad Xilu Wang$^{2}$ \qquad Lu Yin$^{2 \;\dagger}$\thanks{{$^\dagger$} Corresponding author.}}
  
\address{$^{1}$ Institut Polytechnique de Paris, Palaiseau, Île-de-France, France \\
      $^{2}$ University of Surrey, Guildford, Surrey, United Kingdom}

\begin{document}
%
\maketitle
\begin{abstract}
Large Language Models (LLMs) often generate unnecessarily verbose Chain-of-Thought (CoT) reasoning that increases computational costs and latency without proportional performance gains. In this paper, we propose \textbf{F}ine-grained \textbf{G}roup policy \textbf{O}ptimization (\textbf{FGO}), a Reinforcement Learning (RL) algorithm that refines group responses by subdividing them and assigning appropriate weights based on length and entropy, thereby enabling effective CoT compression. Meanwhile, as an enhanced variant of Group Relative Policy Optimization (GRPO), FGO successfully addresses two major limitations of the GRPO: inefficient data utilization and entropy collapse. We evaluate FGO on multiple reasoning LLMs and benchmarks, including MATH500, AIME24, AMC23, and Minerva. Experimental results show that FGO achieves efficient CoT compression without degrading performance, and simultaneously resolves the key limitations of GRPO. Code: \href{https://github.com/Mr-XcHan/FGO}{https://github.com/Mr-XcHan/FGO}.

\end{abstract}
\begin{keywords}
LLM, CoT Compression, RL, GRPO
\end{keywords}
\vspace{-0.2cm}
\section{Introduction}
\label{sec:intro}
\vspace{-0.1cm}

Advances in LLMs have highlighted the central role of CoT reasoning \cite{CoT, Surveyreasoning}, particularly in complex domains such as mathematics and code generation. The Long-CoT reasoning capabilities of LLMs, as exemplified by OpenAI-o1\cite{OpenAI-O1}, DeepSeek-R1\cite{Deepseek-r1}, predominantly stem from RL \cite{ProjIQL, ELAPSE, COSIQL, LeakyPPO} post-training paradigms, which guide models to produce explicit and logically structured reasoning trajectories. However, recent research \cite{O1-pruner, CoTValve} has demonstrated that reasoning ability does not scale linearly with the length of CoT. On the contrary, excessively long CoT often leads to performance degradation due to overthinking and redundant double-checking. This makes it crucial to develop methods that can compress CoT while preserving reasoning performance.

Existing approaches to CoT compression fall into three categories: token-level, instance-level and chunk-level compression. Token-level compression \cite{Tokenskip} reduces length by filtering unimportant tokens but often undermines logical consistency. Instance-level method (e.g.,\cite{C3OT, CoTValve}) relies on an additional compressor LLM, making the performance highly dependent on the auxiliary model. Chunk-level compression \cite{R1-Compress} preserves self-reflection but incurs substantial computational overhead from repeated segmentation and search.

To address these limitations, we propose FGO, a RL-based approach tailored for long CoT compression. FGO extends GRPO, a simple yet efficient RL post-training method, while simultaneously addressing two inherent limitations of GRPO: inefficient data utilization and entropy collapse \cite{DAPO}. Specifically, FGO subdivides responses into correct and incorrect subgroups, and further refines reward assignment by incorporating length and entropy information. This reward shaping discourages unnecessary overthinking without impairing performance and self-reflection (self-reflection demonstrates the model’s ability to evaluate and revise its reasoning process). Moreover, FGO achieves full data utilization while avoiding both entropy collapse.

We evaluate FGO on multiple LLMs and benchmarks, including MATH500, AIME24, AMC23, and Minerva. Results show that FGO achieves substantial CoT compression while maintaining or even improving accuracy. Self-reflection experiments further confirm that compression does not compromise reasoning, and comparison and ablation studies verify FGO’s effectiveness in addressing GRPO’s limitations.

The contributions can be summarized as follows:
\begin{itemize}
    \item We propose FGO, an algorithm that effectively compresses long CoTs while preserving performance.
    \vspace{-0.1cm}
    \item FGO addresses inefficient data utilization and entropy collapse limitations of GRPO through subgrouping and fine-grained reward assignment.
    \vspace{-0.1cm}
    \item Extensive experiments across models and benchmarks demonstrate both the efficiency of CoT compression and the effectiveness of FGO.
\end{itemize}

\begin{figure*}[htbp] 
\centerline{\includegraphics[width=0.9\textwidth]{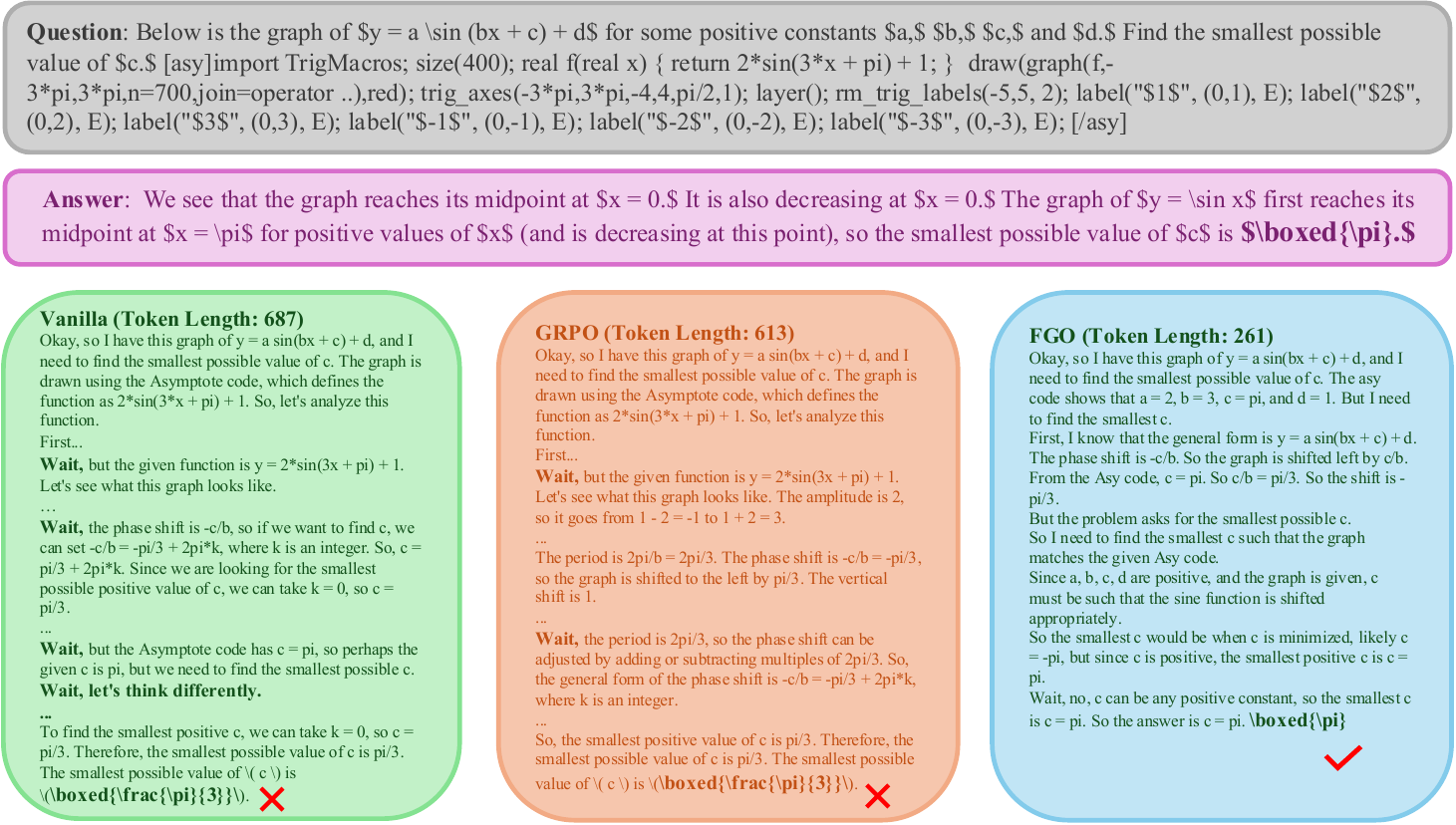}}
\caption{A case study with ZR1-1.5B on MATH500 dataset, comparing Vanilla, GRPO and FGO methods.}
\label{Fig-Example}
\end{figure*}

\vspace{-0.4cm}
\section{Preliminaries}
\label{sec:Preliminaries}
\vspace{-0.3cm}
GRPO estimates the advantage function in the group-relative method. Given a question $s$ and a dataset answer $a$, the old policy $\pi_{\theta_{old}}$ samples a group of $G$ responses $\{o_i\}_{i=1}^G$. The advantage function of the $i$-th response is defined as
\begin{equation}  
\begin{aligned}
    A_{i,t} = \frac{r_i - \text{mean}(\{r_i\}_{i=1 }^{G})}{\text{std}(\{r_i\}_{i=1 }^{G})}.
\label{Eq_GRPO_Adv}
\end{aligned}
\end{equation}

Then, GRPO adopts a clipped objective,
\begin{equation}  
\small
\begin{aligned}
\mathcal{J}_\text{GRPO}(\theta) & =  \mathbb{E}_{(s,a)\sim \mathcal{D}, \{o_i\}_{i=1}^G\sim \pi_{\theta_{old}}(\cdot\mid s)} 
\Bigg[ \frac{1}{G}\sum_{i=1}^{G} \frac{1}{|o_i|}\sum_{t=1}^{|o_i|} \Bigg( \\&
\min \Big( \rho_{i,t}(\theta) A_{i,t},  
\ \text{clip} \Big( \rho_{i,t}(\theta), 1 - \varepsilon, 1 + \varepsilon \Big) A_{i,t} \Big)
\\& - \gamma D_{\text{KL}}(\pi_{\theta} || \pi_{\text{ref}}) 
\Bigg) \Bigg],
\label{Eq_GRPO_obj}
\end{aligned}
\end{equation}
where $\rho_{i,t}(\theta) = \frac{\pi_{\theta}(o_{i,t} \mid s, o_{i,<t})}{\pi_{\theta_{old}}(o_{i,t} \mid s, o_{i,<t})}.$

\textbf{Inefficient data utilization}. When all responses in a group receive identical rewards, the $A_{i,t}$ is zero.

\textbf{Entropy collapse}. During training, the response entropy decreases sharply, leading to nearly identical responses and exacerbates the first issue: inefficient data utilization.

\begin{table*}[!t]
\centering
\small
\setlength{\tabcolsep}{5pt}
\begin{tabular}{ll ||ccc||ccc||ccc||ccc}
\toprule
\multirow{2}{*}{Model} & \multirow{2}{*}{Algo}
 & \multicolumn{3}{c||}{Math500}
 & \multicolumn{3}{c||}{AIME24}
 & \multicolumn{3}{c||}{AMC23}
 & \multicolumn{3}{c}{Minerva} \\
 &  & Acc & Length & ACT 
    & Acc & Length & ACT
    & Acc & Length & ACT
    & Acc & Length & ACT\\
\midrule
\multirow{4}{*}{\shortstack{Qwen2.5-\\Math-1.5B}}
 & Vanilla  & 40.0 & 763 & 5.2  & 10.0 & 1283 & 0.8 & 30.0 & 978 & 3.1 & 9.2  & 1357 & 0.7 \\
 & GRPO     & 65.6 & 578 & 11.3 & 13.6 & 1017 & 1.3 & 50.0 & 825 & 6.1 & 11.4 & 923  & 1.2 \\
 & TLDR     & 68.2 & 686 & 9.9  & -    & -    & -   & -    & -   & -   & -    & -    & -    \\
 & FGO      & \textbf{68.6} & \textbf{441}  & \textbf{15.6} & \textbf{19.3} & \textbf{946}  & \textbf{2.0} & \textbf{52.5} & \textbf{686} & \textbf{7.7} & \textbf{18.0} & \textbf{673}  & \textbf{2.7} \\
\midrule
\multirow{4}{*}{\shortstack{DeepSeek-\\R1-Distill-\\Qwen-1.5B}}
 & Vanilla  & 32.4 & 982 & 3.3  & 3.7  & 1889 & 0.2 & 27.5 & 1820 & 1.5 & 12.1 & 1404 & 0.9 \\
 & GRPO     & 51.0 & 828 & 6.2  & 7.7  & 1879 & 0.4 & 30.0 & 1646 & 1.8 & \textbf{20.6} & 1132 & 1.8 \\
 & TLDR     & 54.6 & 308 & 17.7 & 0.0  & \textbf{494}  & 0   & 19.3 & \textbf{444}  & 4.3 & -    & -    & -    \\
 & FGO      & \textbf{56.4} & \textbf{229} & \textbf{24.6} & \textbf{8.3} & 962  & \textbf{0.9} & \textbf{40.0} & 651  & \textbf{6.1} & 18.0 & \textbf{202}  & \textbf{8.9} \\
\midrule
\multirow{4}{*}{\shortstack{ZR1-1.5B}}
 & Vanilla  & 58.4 & 704 & 8.3  & 13.0  & 1840 & 0.7 & 50.0 & 1450 & 3.4 & 17.3 & 1235 & 1.4 \\
 & GRPO     & 64.0 & 706 & 9.1  & 13.3  & 1840 & 0.7 & 50.0 & 1453 & 3.4 & 23.2 & 1209 & 1.9 \\
 & TLDR     & -    & -   & -    & -     & -    & -   & -    & -    & -   & -    & -    & -    \\
 & FGO    & \textbf{69.0} & \textbf{321} & \textbf{21.5} & \textbf{13.7}  & \textbf{1151} & \textbf{1.2} & \textbf{55.0} & \textbf{540}  & \textbf{10.2}& \textbf{24.6} & \textbf{412}  & \textbf{6.0} \\
\midrule
\multirow{4}{*}{\shortstack{Qwen2.5-\\Math-1.5B-\\Instruct}}
 & Vanilla  & 71.2 & 542 & 13.1 & 12.7  & 890  & 1.4 & 47.5 & 802 & 5.9 & 20.2 & 652  & 3.1 \\
 & GRPO     & 73.0 & 541 & 13.5 & \textbf{14.0}  & 998  & 1.4 & 52.5 & 746  & 7.0 & 22.1 & 645  & 3.4 \\
 & TLDR     & 69.7 & 603 & 11.6 & 7.0   & 1031    & 0.7   & 37.3    & 861   & 4.3   & -    & -    & -    \\
 & FGO      & \textbf{73.2} & \textbf{321} & \textbf{22.8} & 13.3  & \textbf{810}  & \textbf{1.6} & \textbf{55.0} & \textbf{494}  & \textbf{11.1}& \textbf{23.2} & \textbf{374}  & \textbf{6.2} \\
\bottomrule
\end{tabular}
\caption{Comparative results of Vanilla, GRPO and FGO, where bold indicates the best performance.}
\label{Tab-Evaluation}
\end{table*}
\vspace{-0.3cm}
\section{Methodology}
\label{sec:methodology}

Given a question $s$, LLM generates a group of answers $\mathcal{G} = \{\hat{a}_1, \hat{a}_2, ..., \hat{a}_G\}$ extracted from the responses. Each answer $\hat{a}_i$ is compared against the ground-truth answer $a$ in the dataset and assigned a verified reward $r_i$, i.e. 
\begin{equation}  
\begin{cases}
    r_i = 1, \;\;\; & if\;\; \hat{a}_i = a; \\
    r_i = 0, \;\;\; & if\;\; \hat{a}_i \neq a.
\label{Eq-VerifiedR}
\end{cases}
\end{equation}

In GRPO, the reward Eq.~\eqref{Eq-VerifiedR} is directly used for advantage estimation. In contrast, FGO regroups the $\hat{a}_i$ by their verified reward $r_i$ into a correct-response subgroup $\mathcal{G}^+ = \{\hat{a}_i|r_i=1\}$ and an incorrect-response subgroup $\mathcal{G}^- = \{\hat{a}_i|r_i=0\}$, followed by subgroup-specific reward shaping.

For the correct-response subgroup $\mathcal{G}^+$, we retain the reward $\Bar{R}^+ = \{r^+_1=1, r^+_2=1, ..., r^+_{|\mathcal{G}^+|}=1 \}$ to improve the accuracy. Meanwhile, for Long-CoT compression, shorter and more confident (i.e., lower-entropy) responses should receive greater emphasis. Here, response length $L^+ = \{l_1^+, l_2^+, ..., l_{|\mathcal{G}^+|} ^+\}$ is measured by the number of tokens, while confidence is quantified by the entropy $\mathcal{H}^+ = \{\mathcal{H}_1^+, \mathcal{H}_2^+, ..., \mathcal{H}_{|\mathcal{G}^+|} ^+\}$, where
\begin{equation}  
\begin{aligned}
\mathcal{H}_{i} = \sum_{t=1}^{|o_i|} \left[ - \mathbb{E}_{o_{i,t} \sim \pi_{\theta}(\cdot|s, o_{i, <t})} \left[log \  \pi_{\theta}(o_{i,t}|s, o_{i, <t}) \right] \right].
\label{Eq-EntropyDef}
\end{aligned}
\end{equation}

Then, the fine-grained weight $\mathcal{W}^+$ is computed as
\begin{equation}  
\begin{aligned}
\mathcal{W}^+ = \text{Softmax} \left[ \Big( \frac{ \text{mean} (L^+)}{L^+} \Big)^\alpha \times \Big( \frac{  \text{mean}(\mathcal{H}^+)}{\mathcal{H}^+} \Big) ^ \beta \right].
\label{Eq-weight+}
\end{aligned}
\end{equation}

In Eq.~\eqref{Eq-weight+}, $L^+$ and $\mathcal{H}^+$ in the denominator indicate that shorter responses and lower-entropy responses are assigned larger weights $\mathcal{W}^+$. To avoid scale sensitivity, both response length and entropy are normalized using $\text{mean}(L^+)$ and $\text{mean}(\mathcal{H}^+)$. The hyperparameter $\alpha$ controls the degree of CoT length compression, specifically, larger $\alpha$ values encourage shorter CoT responses. $\beta$ controls the degree of exploration. In addition, a softmax operator is applied to normalize $\mathcal{W}^+$ within the correct-response subgroup. Finally, the fine-grained reward is defined as
\begin{equation}  
\begin{aligned}
R^+ = \mathcal{W}^+ \times \Bar{R}^+. 
\label{Eq-R+}
\end{aligned}
\end{equation}

For incorrect-response subgroup $\mathcal{G}^-$, verified rewards are modified from $0$ to $-1$, i.e., $\Bar{R}^- = \{r^-_1=-1, r^-_2=-1, ..., r^-_{|\mathcal{G}^-|}=-1\}$, penalizing incorrect answers. This adjustment is necessary because when $r^-_i=0$, multiplying by any weight $w^-_i$ renders $w^-_i = 0$ ineffective. Unlike $\mathcal{G}^+$, within the incorrect-response subgroup $\mathcal{G}^-$, shorter and more exploratory (i.e., higher-entropy) responses should receive greater weight. Similarly, by defining $L^- = \{l_1^-, l_2^-, ..., l_{|\mathcal{G}^-|} ^-\}$ and $\mathcal{H}^- = \{\mathcal{H}_1^-, \mathcal{H}_2^-, ..., \mathcal{H}_{|\mathcal{G}^-|} ^-\}$, the weight $\mathcal{W}^-$ is computed as follows:
\vspace{-0.22cm}
\begin{equation}  
\begin{aligned}
\mathcal{W}^- = \text{Softmax} \left[ \Big(\frac{L^-}{\text{mean}(L^-)} \Big)^{\alpha} \times \Big( \frac{\text{mean}(\mathcal{H}^-)}{\mathcal{H}^-} \Big)^{\beta} \right].
\label{Eq-weight-}
\end{aligned}
\end{equation}
\vspace{-0.22cm}

Therefore, the fine-grained reward for $\mathcal{G}^-$ is calculated by
\vspace{-0.2cm}
\begin{equation}  
\begin{aligned}
R^- = \mathcal{W}^- \times \Bar{R}^-. 
\label{Eq-R-}
\end{aligned}
\end{equation}
\vspace{-0.3cm}

We construct the complete reward set $R = \{R^+, R^-\}$. The advantage function is computed as Eq.~\eqref{Eq_DrGRPO_Adv},
\vspace{-0.1cm}
\begin{equation}  
\begin{aligned}
    A_{i,t} = r_i - \text{mean}(\{r_i\}_{i=1 }^{G}), \; \; r_i \in R,
\label{Eq_DrGRPO_Adv}
\end{aligned}
\end{equation}
where, following Dr.GRPO \cite{DrGRPO}, the standard deviation term is omitted for stability. Then, the optimization objection remains consistent with Eq.~\eqref{Eq_GRPO_obj}. 

Through subgroup-specific reward shaping, FGO compresses long CoTs by prioritizing shorter responses while preserving accuracy. By integrating both length and entropy signals, FGO consistently achieves a $100\%$ data utilization rate across experiments. Furthermore, the relative entropy optimization between correct and incorrect response groups effectively mitigates entropy collapse.

\section{Experiments}

\subsection{Experiment Settings}
We apply FGO to four reasoning LLMs tailored for mathematics, including Qwen2.5-Math-1.5B \cite{Qwen2.5-Math}, DeepSeek-R1-Distill-Qwen-1.5B \cite{DeepSeek-R1-1.5B}, ZR1-1.5B \cite{ZR1-1.5B}, and Qwen2.5-Math-1.5B-Instruct \cite{Qwen2.5-Math-Ins}. The training set consists of the first $3,200$ samples from MATH-Lighteval \cite{MathLighteval}. We adopt prompt format in \cite{O1-pruner}: \emph{``You are a helpful assistant. You should think step-by-step and put your final answer within boxed \{\{\}\}."}. We fix $\gamma=0$ in Eq.~\eqref{Eq_GRPO_obj}, i.e., without reference model alignment, to reduce memory and computational overhead. We set $\alpha=0.01, \beta=1$ as the default and conduct ablation experiments to analyze the effect of varying $\alpha$ and $\beta$.
\vspace{-0.3cm}

\subsection{Main Results}
The trained models are evaluated on four benchmarks, with results summarized in Tab.~\ref{Tab-Evaluation}. The evaluation metrics are: $(i)$ Accuracy (Acc), measured by \emph{pass@$1$}; $(ii)$ Token-Length (Length), the average number of the generated tokens; $(iii)$ Accuracy Contribution per hundred Tokens (ACT). Since the AIME24 contains relatively few samples, we report \emph{avg@$10$} accuracy to mitigate randomness. For TLDR \cite{TLDR}, we adopt the results reported in the original paper.

As shown in Tab.~\ref{Tab-Evaluation}, FGO leads to a substantial reduction in CoT length. In addition, FGO achieves higher Acc metric. Fig.~\ref{Fig-Example} provides a representative example with the ZR1-1.5B on the $17th$ problem from MATH500. Although Vanilla and GRPO produce longer CoT by double-checking, the final answers remain incorrect. The ACT metric further reflects that FGO achieves more efficient token utilization compared with the GRPO and TLDR methods.

\begin{table*}[!t]
\centering
\small
\setlength{\tabcolsep}{5pt}
\begin{tabular}{ll ||ccc||ccc||ccc||ccc}
\toprule
\multirow{2}{*}{Model} & \multirow{2}{*}{FGO}
 & \multicolumn{3}{c||}{Math500}
 & \multicolumn{3}{c||}{AIME24}
 & \multicolumn{3}{c||}{AMC23}
 & \multicolumn{3}{c}{Minerva} \\
 &  & Acc & Length & ACT 
    & Acc & Length & ACT
    & Acc & Length & ACT
    & Acc & Length & ACT\\
\midrule
\multirow{3}{*}{\shortstack{Qwen2.5-\\Math-1.5B}}
 & $\alpha=1$   & 67.0 & \textbf{342}  & \textbf{19.6} & 10.0 & \textbf{740}  & 1.4 & 40.0 & \textbf{578} & 6.9 & \textbf{20.6} & \textbf{662}  & \textbf{3.1} \\
 & $\alpha=0.01$& \textbf{68.6} & 441  & 15.6 & \textbf{19.3} & 946  & \textbf{2.0} & \textbf{52.5} & 686 & \textbf{7.7} & 18.0 & 673  & 2.7 \\
 & $\alpha=0$   & 67.6 & 437  & 15.5 & 16.0 & 959  & 1.7 & 45.0 & 708 & 6.4 & 18.4 & 695  & 2.6 \\
\midrule
\multirow{3}{*}{\shortstack{DeepSeek-\\R1-Distill-\\Qwen-1.5B}}
 & $\alpha=1$   & 44.4 & 340  & 13.1 & 6.7  & 1450 & 0.5 & 32.5 & \textbf{623}  & 5.2 & 18.0 & 302  & 6.0 \\
 & $\alpha=0.01$& \textbf{56.4} & \textbf{229}  & \textbf{24.6} & \textbf{8.3}  & \textbf{962}  & \textbf{0.9} & \textbf{40.0} & 651  & \textbf{6.1} & 18.0 & \textbf{202}  & \textbf{8.9} \\
 & $\alpha=0$   & \textbf{56.4} & 484  & 11.7  & 6.7  & 1290 & 0.5 & 35.0 & 831  & 4.2 & \textbf{18.8} & 490  & 3.8 \\
\midrule
\multirow{3}{*}{\shortstack{ZR1-1.5B}}
 & $\alpha=1$   & 67.6 & \textbf{283} & \textbf{23.9} & 10.0  & \textbf{973}  & 1.0 & 52.5 & 568  & 9.2 & 20.6 & \textbf{373}  & 5.5 \\
 & $\alpha=0.01$& 69.0 & 321 & 21.5 & 13.7  & 1151 & 1.2 & 55.0 & \textbf{540}  & \textbf{10.2}& \textbf{24.6} & 412  & \textbf{6.0} \\
 & $\alpha=0$   & \textbf{71.8} & 456 & 15.7 & \textbf{23.0}  & 1517 & \textbf{1.5} & \textbf{60.0} & 844  & 7.1 & 23.2 & 629  & 3.7 \\
\midrule
\multirow{3}{*}{\shortstack{Qwen2.5-\\Math-1.5B-\\Instruct}}
 & $\alpha=1$   & 68.0 & \textbf{290} & \textbf{23.4} & 10.0  & \textbf{565}  & \textbf{1.8} & 50.0 & \textbf{447}  & \textbf{11.2}& 22.4 & \textbf{354}  & \textbf{6.3} \\
 & $\alpha=0.01$& \textbf{73.2} & 321 & 22.8 & \textbf{13.3}  & 810  & 1.6 & \textbf{55.0} & 494  & 11.1& \textbf{23.2} & 374  & 6.2 \\
 & $\alpha=0$   & 70.0 & 440 & 15.9 & 6.3   & 801  & 0.8 & 52.5 & 631  & 8.3 & 20.6 & 498  & 4.1 \\
\bottomrule
\end{tabular}
\caption{FGO with $\alpha=1, 0.01, 0$ ablation results, where bold indicates the best performance.}
\label{Tab-Ablation}
\end{table*}

\subsection{Self-Reflection Results}
\begin{figure}[htbp] 
\centerline{\includegraphics[width=0.5\textwidth]{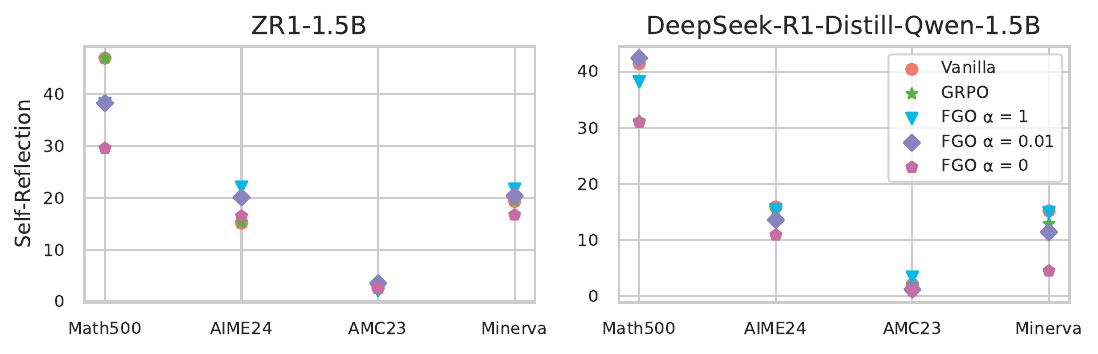}}
\vspace{-0.2cm}
\caption{The Self-Reflection keywords count.}
\label{Fig-Self-Reflection}
\end{figure}

As self-reflection is a crucial capability within Long-CoT. Therefore, we conduct experiments to examine the self-reflection ability of the models trained by FGO. We count the occurrences of eflection-related keywords \cite{R1-Compress}: ``wait", ``alternatively", ``emm", ``hmm" per hundred tokens. As illustrated by the results in Fig.~\ref{Fig-Self-Reflection}, FGO preserves the majority of self-reflection steps and does not lose reasoning capability despite the compression of CoT length.
\vspace{-0.2cm}
\vspace{-0.5cm}

\subsection{Results on Eliminating the Two Limitations of GRPO}
When all responses within a group receive identical rewards, GRPO degenerates. Tab.~\ref{Tab-DataUtilization} reports the number of such cases among the $3,200$ training samples, clearly indicating that GRPO suffers from inefficient data utilization, whereas FGO consistently achieves $100\%$ utilization.
\vspace{-0.2cm}
\begin{table}[htbp]
\setlength{\tabcolsep}{4pt}
\small
\begin{tabular}{c|cccc}
\toprule
     & \footnotesize{\shortstack{Qwen2.5-\\Math-1.5B}} & \footnotesize\shortstack{DeepSeek-R1-\\Distill-Qwen-1.5B} & \footnotesize{ZR1-1.5B} & \footnotesize\shortstack{Qwen2.5-Math-\\1.5B-Instruct} \\
\midrule
GRPO & 1584              & 1866                          & 2617     & 2444\\
\midrule
FGO  & 0                 & 0                             & 0        & 0 \\ \bottomrule
\end{tabular}
\caption{Number of invalid samples for GRPO and FGO.}
\label{Tab-DataUtilization}
\end{table}
\vspace{-0.2cm}

\begin{figure}[htbp] 
\centerline{\includegraphics[width=0.5\textwidth, height=0.15\textwidth]{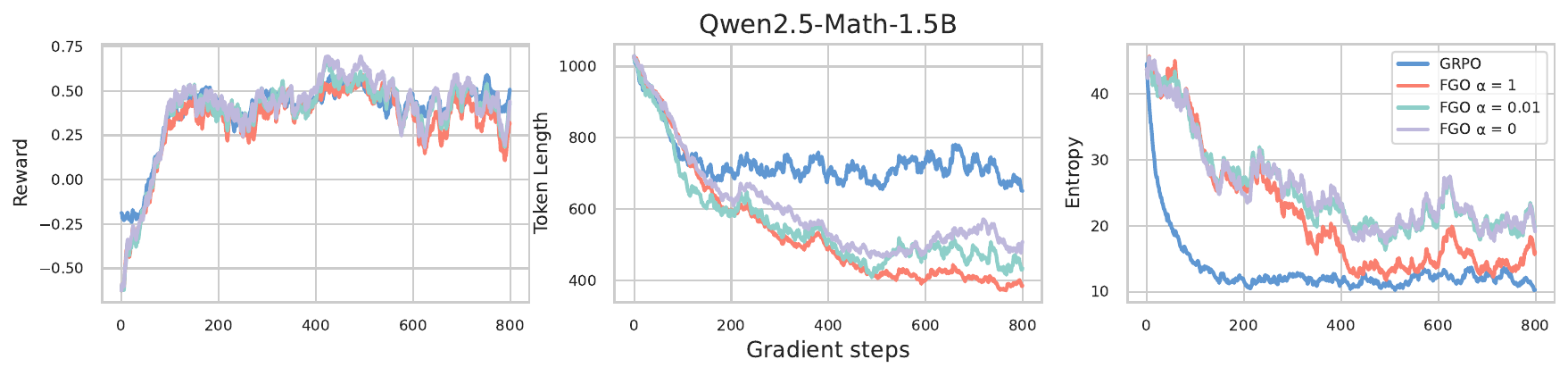}}
\vspace{-0.2cm}
\caption{GRPO and FGO training curves on Qwen2.5-Math-1.5B Model, including reward, length and entropy.}
\label{Fig-Qwen-Training}
\end{figure}

Another limitation of GRPO is entropy collapse. Fig.~\ref{Fig-Qwen-Training} shows the trajectory-level entropy dynamics of Qwen-Math-1.5B during training, together with reward and token-length curves. Under FGO, entropy decreases more gradually and ultimately remains higher, demonstrating that FGO effectively mitigates entropy collapse and preserves sufficient exploration.
\vspace{-0.2cm}
\vspace{-0.4cm}

\subsection{Ablation Experiments}
An important hyperparameter in FGO is $\alpha$, which controls response length. We evaluate three settings, $\alpha \in \{1, 0.01, 0\}, \beta = 1$, with results summarized in Tab.~\ref{Tab-Ablation}.

As shown in Tab.~\ref{Tab-Ablation}, larger $\alpha$ produces shorter responses but yields a lower ACC, while $\alpha=0$ results in longer responses yet still fails to achieve the best accuracy, consistent with the reasoning ability does not scale linearly with the length of CoT. In contrast, $\alpha=0.01$ achieves the best overall performance in both ACC and ACT.

\section{Conclusion}
\label{sec:conclusion}

In this work, we propose the FGO algorithm, which refines group-based responses to effectively compress CoT length while maintaining, or even improving, LLM performance. As an enhanced variant of GRPO, FGO addresses two critical limitations: inefficient data utilization and entropy collapse. Extensive experiments validate the effectiveness of FGO. Future work may explore strategies for estimating more accurate advantage functions with fewer group responses.

\newpage
\bibliographystyle{IEEEbib}
\bibliography{refs}

\end{document}